\documentclass{article}
\usepackage[numbers]{natbib}
\usepackage{amsmath}
\usepackage{tikz}
\usepackage{float}
\usetikzlibrary{shapes.geometric, arrows.meta, positioning, calc}
\usepackage{listings}
\usepackage{xcolor}

\usepackage[preprint]{neurips_2025}

\usepackage[utf8]{inputenc} 
\usepackage[T1]{fontenc}    
\usepackage{hyperref}       
\usepackage{url}            
\usepackage{booktabs}       
\usepackage{amsfonts}       
\usepackage{nicefrac}       
\usepackage{microtype}      
\usepackage{xcolor}         

\title{Synthesizing Procedural Memory: Challenges and Architectures in Automated Workflow Generation}

\author{%
  Nishant Gaurav \\
  \texttt{nishant@agentr.dev} \\
  \And
  Adit Akarsh \\
  \texttt{adit@agentr.dev} \\
  \AND
  Ankit Ranjan \\
  \texttt{ankit@agentr.dev} \\
  \And
  Manoj Bajaj \\
  \texttt{manoj@agentr.dev} \\
}

\begin{document}

\maketitle

\begin{abstract}
While \textit{CodeMem} \cite{gaurav2025codemem} establishes executable code as the optimal representation for agentic procedural memory, the mechanism for autonomously synthesizing this memory from a ``blank slate'' remains underexplored. This paper operationalizes the transition of Large Language Models (LLMs) from passive tool-users to active workflow architects. Through a high-fidelity case study of a cross-service orchestration task (Outlook to OneDrive), we identify and address four structural bottlenecks in automated skill generation: the \textit{Discovery Gap} (navigating large tool registries via Dynamic MCP), the \textit{Verification Gap} (grounding tool response structures via a ``Probe'' methodology), the \textit{Decomposition Gap} (replacing inefficient inference-time search with Linear State Anchoring), and the \textit{Scaling Gap} (architecting for concurrency and persistence). We demonstrate that by enforcing a scientific methodology-- hypothesize, probe, and code--agents can autonomously write robust, production-grade code skills.
\end{abstract}

\section{Introduction}

Recent work on code-based agentic architectures establishes that the optimal representation of agentic procedural memory is not natural language instructions, but executable, deterministic code \cite{gaurav2025codemem, wang2024executable}. While this framework defines the final form as a repository of frozen, reliable skills, it does leave open the question of their formation. How does a probabilistic model blindly navigate an unknown toolset to construct a deterministic program? 

The transition from a passive chatbot to an automated code writer is not straightforward. An agent cannot simply ``write code'' for an API it has never seen, nor can it process data structures it has not verified \cite{chen2021codex}. To successfully bridge the gap between user intent and executable logic, the agent must adopt a scientific methodology: hypothesizing tool utility, conducting experiments to probe data shapes, and iteratively refining logic against feedback \cite{sumers2024coala}. 

To ground our analysis in practical reality, we utilize a running example of a complex administrative workflow throughout this paper: ``Scan the last 15 days of Outlook emails for PDF or XLSX attachments, filter out internal communications (e.g., from @agentr.dev), and archive the remaining files into dynamic OneDrive folders based on the sender’s company name.'' This seemingly routine task exposes the fragility of standard approaches, requiring precise handling of API discovery, data validation, and state management. 

We dissect the four fundamental bottlenecks of automated workflow synthesis required to solve such a task: 

\begin{enumerate}
    \item \textbf{The Discovery Gap:} Identifying relevant capabilities (e.g., Outlook vs. Gmail) in an infinite tool registry, a challenge exacerbated in large-scale environments \cite{gaurav2025dynamic}. 
    \item \textbf{The Verification Gap:} Grounding tool response schemas (e.g., attachment metadata structure) through active sampling to avoid hallucinated data access \cite{nguyen2024dynasaur}. 
    \item \textbf{The Decomposition Gap:} Structuring logic (e.g., loops and conditionals) using intrinsic reasoning \cite{chen2022program} rather than inefficient and computationally expensive search trees \cite{zhou2023lats}. 
    \item \textbf{The Scaling Gap:} Architecting for concurrency and persistence (e.g., processing 5,000 emails) in production environments, necessitating robust memory frameworks \cite{langchain2024persistence}.
\end{enumerate}

We propose specific structural requirements and interaction patterns to resolve these challenges, enabling agents to autonomously author robust, high-utility skills that reside within a persistent procedural memory \cite{langchain_memory_concepts}.
\section{The Discovery Gap: Dynamic Tool Identification}

The first bottleneck in synthesizing procedural memory for an agent in a large MCP environment is figuring out its capabilities, i.e., which applications it can use and how. When an agent is initialized to perform a task like ``Analyze my last 50 emails,'' it possesses a goal but lacks the means. It does not inherently know if the available environment supports Outlook, Gmail, or IMAP, nor does it know the specific function signatures (e.g., \texttt{outlook\_list\_messages} vs. \texttt{outlook\_list\_emails}). 

\subsection{The Context-Scale Trade-off}

Standard agent architectures attempt to solve this by injecting the entire tool definition list into the system prompt. 

\begin{itemize}
    \item \textbf{Small Scale (10 Tools):} Feasible. The agent sees all capabilities. 
    \item \textbf{Enterprise Scale (1000+ Tools):} Impossible. Injecting thousands of JSON schemas exceeds context windows and degrades model attention, leading to ``function hallucination'' where the model invents convenient but non-existent parameters \cite{chen2021codex}. 
\end{itemize}

To scale to enterprise-grade tool registries, we must transition from \textit{Passive Context Loading} ($O(N)$ cost) to \textit{Active Information Retrieval} ($O(1)$ cost) \cite{gaurav2025dynamic}.

\subsection{Mechanism: Semantic Intent to Registry Query}
We implement a \textbf{Dynamic Model Context Protocol (MCP)} system, as implemented in \cite{gaurav2025dynamic}. Instead of loading tools/functions at start-up, the agent is equipped with a pair of meta-tools: \texttt{search\_functions and load\_functions}, along with a list of the application names (without the full list of functionality). This forces the agent to explicitly formulate a search strategy before attempting execution.

In our running example, the interaction log demonstrates this active discovery process:

\begin{enumerate}
    \item \textbf{Intent Extraction:} The agent parses the user's request ("Outlook," "Attachments," "OneDrive").
    \item \textbf{Registry Query:} Instead of hallucinating function names, the agent issues a semantic search:
\begin{lstlisting}
requests:
  - query: fetch emails download attachment
    app_id: outlook
  - query: create folder upload file list items
    app_id: onedrive
\end{lstlisting}
    \item \textbf{Schema Injection (Just-in-Time):} The system retrieves the top $k$ relevant schemas (e.g., \texttt{outlook\_\_list\_emails}, \texttt{onedrive\_\_upload\_file}) and injects \textit{only} those into the active context via \texttt{load\_functions}.
\end{enumerate}

\subsection{Iterative Refinement}
Crucially, the discovery process is not always single-shot. As seen in the interaction logs, the agent's initial search for "fetch emails" might return only attachment-specific tools. A capable agent must recognize this gap. In our trace, the agent explicitly reasons: \textit{"I'm broadening the search terms to 'list messages' and 'get emails' to see if that uncovers the desired functionality."}

This \textbf{Search $\rightarrow$ Evaluate $\rightarrow$ Refine} loop is the foundational step of synthesizing procedural memory. It ensures that the resulting script is built upon verified, existing capabilities rather than probabilistic guesses.

\subsection{The Necessity of Semantic Search}
Injecting all available tool schemas into the system prompt is computationally prohibitive ($O(N)$ context growth). We propose a retrieval-based discovery mechanism:
\begin{enumerate}
    \item \textbf{Intent Extraction:} The agent parses the user query for semantic domains (e.g., "communication", "file storage").
    \item \textbf{Registry Query:} The agent invokes a \texttt{search\_functions} tool to query a vector database of tool descriptions.
    \item \textbf{Schema Injection:} Only the relevant function definitions (e.g., \texttt{outlook\_\_get\_messages}), called using \texttt{load\_functions} are loaded into the context.
\end{enumerate}
This transforms tool selection from a passive context-reading task into an active information-retrieval task, enabling agents to operate across infinite tool libraries.

\section{The Verification Gap: The "Probe" Methodology}

Even when the correct tool is discovered (e.g., \texttt{outlook\_\_list\_emails}), a critical failure mode remains: \textit{Structure Hallucination}. Large Language Models are probabilistic token predictors; when writing code to handle API responses, they often "guess" the JSON schema based on training data rather than reality.

\subsection{The Hallucination of Shape}
A common error pattern is the assumption of simplified data structures. For example, an LLM might assume an email API returns a direct list of objects:
\begin{lstlisting}[language=python]
# Hallucinated Logic
emails = response  # Assumes list
for email in emails:
    print(email['subject'])
\end{lstlisting}
However, enterprise APIs (like Microsoft Graph) often wrap data in metadata envelopes (e.g., \texttt{\{ "@odata.context": "...", "value": [...] \}}). Without verification, the generated code fails immediately with \texttt{TypeError: list indices must be integers}, causing the agent to crash.

\subsection{The Sampling Mandate}
To bridge this gap, CodeMem enforces a \textbf{Probe Phase}. Before authoring the final, scalable workflow, the agent must perform a "Sample Run." This is not merely a debugging step; it is a structural requirement for synthesizing reliable procedural memory.

The methodology consists of three atomic steps:
\begin{enumerate}
    \item \textbf{Sample (Limit $k$):} The agent calls the API with strict pagination limits (e.g., \texttt{top=5} or \texttt{limit=1}) to minimize latency and token costs.
    \item \textbf{Inspect (Stdout Grounding):} Instead of silently processing the data, the agent prints the raw structure to the sandbox's standard output.
    \item \textbf{Codify (Schema Locking):} The agent reads the logs to "lock" the schema understanding before writing the final logic.
\end{enumerate}

\subsection{Case Study: Probing the Outlook API}
In our running example, the interaction log demonstrates this discipline. The agent does not attempt to write the complex PDF extraction logic immediately. Instead, it executes an intermediate script:

\begin{lstlisting}[language=python, caption=The Probe Step]
# Actual code from Interaction Log
async def _fetch_and_filter_emails():
      start_date = await _get_start_date_iso()
      # ... fetch logic ...
      print(f"Fetched {len(emails)} emails with attachments.")
      # The agent implicitly validates the 'attachments' key exists here
\end{lstlisting}

By observing the output—\texttt{"Fetched 7 emails with attachments"}—the agent verifies that its date calculation logic (\texttt{datetime.now() - timedelta(days=15)}) matches the API's time format requirements and that the `filter` parameter functioned correctly. Had the output been "0 emails," the agent would have known to debug the date format string (ISO 8601) before building the downstream OneDrive logic. This "Fail Fast" mechanism is crucial for autonomous reliability.

\section{The Decomposition Gap: Intrinsic Reasoning vs. Tree Search}

Once the necessary tools are discovered and the data structures verified via probing, the agent must plan the logic. How does an agent decompose a high-level request like "Organize my attachments" into a sequence of atomic actions? Recent literature emphasizes complex inference-time search algorithms like Language Agent Tree Search (LATS) \cite{zhou2023lats}, which combines Monte Carlo Tree Search (MCTS) with self-reflection.

However, we argue that for modern foundational models applied to linear workflows, these heavy search architectures yield diminishing returns and introduce unnecessary latency.

\subsection{The Saturation of Search Benefits}
The LATS framework demonstrated significant value with weaker models (e.g., GPT-3.5), where the model required external scaffolding to correct its own logic. However, modern models (GPT-5, Gemini 3) demonstrate \textit{intrinsic reasoning} capabilities that render these heavy search trees inefficient.

\subsubsection{The Reflection Trap}
LATS relies on a cycle of generation, feedback, and reflection. In algorithmic tasks, this often creates a "hallucination loop." We analyzed performance on \texttt{HumanEval/83} \cite{chen2021codex} ("Count n-digit numbers starting or ending with 1"). Even when powered by GPT-4, a LATS-based agent detecting a failure often enters a spiral of ineffective error correction:
\begin{itemize}
    \item \textbf{The Loop:} The agent generates dozens of reflections (e.g., "The formula double counts... subtract twice...").
    \item \textbf{The Consequence:} Instead of deriving the principle, the agent patches the code with "magic numbers" based on failed test cases.
    \item \textbf{The Result:} High latency and "spaghetti code."
\end{itemize}

\subsection{The Modern Approach: Intrinsic Derivation}
In contrast, modern high-reasoning models solve the same problem via \textit{intrinsic derivation}. Instead of writing code $\rightarrow$ testing $\rightarrow$ patching, the model utilizes Chain of Thought to apply mathematical principles before a single line of code is written.

\begin{figure}[H]
    \centering
    \begin{minipage}{0.48\textwidth}
        \centering
        \textbf{LATS (GPT-4) / Recursive Reflection}
        \scriptsize
        \begin{verbatim}
Reflections:
1. "Formula should be 2 * 10^(n-1)..."
2. "Wait, subtract 10^(n-2) * 2?"
3. "Maybe handle n=2 separately?"

Final Code (From Search Tree):
def starts_one_ends(n):
    if n == 1: return 1
    # Incorrectly assumes base 9 logic
    starts_with_1 = 9 ** (n - 1)
    ends_with_1 = 8 * (9 ** (n - 2))
    overlap = 9 ** (n - 2)
    return starts_with_1 + ends_with_1 - overlap
        \end{verbatim}
    \end{minipage}\hfill
    \begin{minipage}{0.48\textwidth}
        \centering
        \textbf{Modern Intrinsic Reasoning (GPT-5)}
        \scriptsize
        \begin{verbatim}
Derivation:
1. Start w/ 1: 10^(n-1)
2. End w/ 1: 9 * 10^(n-2)
3. Intersect: 10^(n-2)
4. Union = 10^(n-1) + 9*10^(n-2) - 10^(n-2)
         = 18 * 10^(n-2)

Final Code:
def starts_one_ends(n):
    if n == 1:
        return 1
    return 18 * 10**(n - 2)
        \end{verbatim}
    \end{minipage}
    \caption{Comparative analysis of HumanEval/83. The Search-based agent (Left) gets lost in a "patch loop," while the Intrinsic Reasoning agent (Right) produces a deterministic $O(1)$ solution.}
    \label{fig:lats_vs_modern}
\end{figure}

\subsection{Linear State Anchoring: The \texttt{todos} Tool}
For agentic workflows (unlike math problems), the challenge is not finding a "hidden" solution path via search, but maintaining state across a long sequence of obvious steps. A Search Tree is the wrong data structure for a linear pipeline like our Outlook-OneDrive task.

Instead, we introduce linear State anchoring via a \texttt{write\_todos} tool. This acts as an external Program Counter for the agent. It decouples the \textit{plan} from the \textit{context window}.

In our interaction log, the agent explicitly serializes its decomposition before writing code. Note how it breaks the Outlook task into dependencies:

\begin{lstlisting}[caption=Agent Generates State Anchor for Outlook Task]
todos:
  - status: in_progress
    content: Load functions for Outlook and OneDrive
  - content: Fetch Outlook emails (past 15 days) and filter for (.pdf, .xlsx)
    status: pending
  - status: pending
    content: "Process sample: Download, extract Company Name, determine Document Name"
  - status: pending
    content: Create 'Email Attachments December/{Company Name}' folders and upload
\end{lstlisting}

\subsubsection{Why Anchoring Beats Search}
By committing this plan to the \texttt{todos} object, the agent achieves three things that LATS cannot offer efficiently:
\begin{enumerate}
    \item \textbf{Resilience to Context Shift:} Even if the tool outputs (e.g., massive email JSONs) flush the chat history, the `todos` object remains injected at the top of the context. The agent knows it is on Step 3 ("Process sample") regardless of token displacement.
    \item \textbf{Drift Prevention:} The agent cannot "forget" to create the folders. The `pending` status serves as a forcing function for the next generation step.
    \item \textbf{Modularity:} It allows the agent to pause after the "Process sample" step to verify the schema (The Verification Gap) before committing to the bulk upload step.
\end{enumerate}

\section{The Scaling Gap: From Prototype to Production}

The final challenge in the CodeMem pipeline is the transition from a functional prototype to a production-grade workflow. A script that successfully processes 5 emails in a test run will often fail catastrophically when applied to 5,000 emails due to timeouts, rate limits, or memory overflows.

To bridge this gap, the agent must transition from writing "scripting code" to writing "systems code."

\subsection{Concurrency vs. Sequence}
By default, LLMs operate sequentially—they generate code that reads like a narrative (Step A, then Step B). In data-intensive tasks like our Outlook-OneDrive bridge, this default behavior is inefficient.

\begin{itemize}
    \item \textbf{The Naive Approach (Sequential):}
    \begin{lstlisting}[language=python]
    for email in emails:
        # Blocks for 2s per file
        download_attachment(email) 
    # Total time for 1000 emails: ~33 minutes
    \end{lstlisting}
    
    \item \textbf{The CodeMem Approach (Concurrent):}
    To solve this, the agent must be prompted or trained to utilize \texttt{asyncio}. In the interaction log, we observe the agent defining functions with \texttt{async def}, preparing the architecture for non-blocking I/O.
    \begin{lstlisting}[language=python]
    # Agent generated async architecture
    tasks = [download_attachment(email) for email in emails]
    await asyncio.gather(*tasks)
    # Total time for 1000 emails: ~2 minutes (limited by bandwidth)
    \end{lstlisting}
\end{itemize}

\subsection{External Persistence: The "Memory" Proposal}
For long-running processes, relying on the LLM's context window or the sandbox's RAM to track state is fatal. If the process crashes at email \#4,999, a stateless script must restart from zero, creating thousands of duplicates.

We tackle this via \textbf{External Persistence}—using lightweight databases (SQLite, Google Sheets) as checkpoints.

Crucially, our running example demonstrates the agent's intrinsic awareness of this gap. In the interaction log, before finalizing the agent, the system pauses to ask:

\begin{quote}
\textbf{Agent:} "Would you like me to add a 'memory' (e.g., using a Google Sheet) to track processed emails? ... This avoids duplicates if you run it multiple times."
\end{quote}

Although the user declined ("no log required") for this specific instance, this interaction proves the \textbf{Architect/Builder} model. The agent recognized that the logic was technically correct but operationally fragile without a state store, and proactively proposed a scaling solution.

\subsection{Idempotency at the Edge}
Even without a database log, the agent implemented \textbf{Idempotency} at the infrastructure level. In the final script, instead of blindly creating folders, it implemented logic to handle existing artifacts:
\begin{lstlisting}[language=python]
try:
    await onedrive__create_folder(name=folder_name)
except:
    pass # If folder exists, proceed to upload
\end{lstlisting}
This defensive coding ensures that re-running the workflow on the same dataset does not result in errors, effectively solving the reliability problem for the "No Log" constraint.

\section{Conclusion}

The transition from Chat-based Tool Calling to Code-based Workflow Synthesis represents a fundamental maturation in agentic AI. As we have demonstrated through the Outlook-OneDrive case study, the ability to execute a task is distinct from the ability to architect a reliable solution for it.

This paper has defined the implementation pipeline required to bridge the gap between ephemeral reasoning and persistent procedural memory. We identified that:
\begin{itemize}
    \item \textbf{Discovery} must move from passive context loading to active information retrieval via Dynamic MCP.
    \item \textbf{Verification} requires a "Probe Phase" to ground hallucinated schemas in observed reality.
    \item \textbf{Decomposition} is best achieved through Linear State Anchoring (\texttt{write\_todos}) and intrinsic reasoning, rejecting the diminishing returns of heavy search trees like LATS for linear tasks.
    \item \textbf{Scaling} demands that agents be prompted to act as systems engineers, utilizing concurrency and external state to ensure production-grade reliability.
\end{itemize}

By solving these four structural bottlenecks—Discovery, Verification, Decomposition, and Scaling—we empower Large Language Models to graduate from being passive routers of API calls to active authors of their own capabilities. The result is a coding agent that does not merely answer questions, but builds the verifiable, reproducible tools necessary to answer them at scale.

\bibliography{main}

\end{document}